\documentclass[letterpaper, 10 pt, conference]{ieeeconf} 

\providecommand{\citet}[1]{\cite{#1}}    %
\providecommand{\citep}[1]{\cite{#1}}    %

\usepackage{cite}
\usepackage[colorlinks,urlcolor=blue!70!black]{hyperref}

\usepackage{amsmath,amsfonts,bm}

\def\eqref#1{equation~\ref{#1}}

\def\1{\bm{1}}

\DeclareMathAlphabet{\mathsfit}{\encodingdefault}{\sfdefault}{m}{sl}
\SetMathAlphabet{\mathsfit}{bold}{\encodingdefault}{\sfdefault}{bx}{n}

\usepackage{amssymb}
\let\labelindent\relax
\usepackage{enumitem}
\usepackage{xspace}
\usepackage{booktabs}
\usepackage{colortbl}
\usepackage{pifont}
\usepackage{graphicx}
\usepackage{siunitx}
\usepackage{cleveref}
\usepackage{makecell}
\usepackage{multirow}
\usepackage[table,dvipsnames]{xcolor} %

\usepackage{algorithm}
\usepackage{algpseudocode}

\def\llara{LLaRA\xspace}
\def\modelname{LLaRA+IVRA\xspace}

\definecolor{Gray}{gray}{0.90}
\definecolor{lightblue}{rgb}{0.9,0.96,1}

\newcommand{\inc}[1]{\ensuremath{_{\text{\textcolor{PineGreen}{(+#1)}}}}}

\usepackage{subcaption} %

\definecolor{myblue}{RGB}{235,245,251}
\definecolor{down}{RGB}{153, 163, 164}
\definecolor{down2}{RGB}{153, 204, 205}

\usepackage{wrapfig}

\usepackage{calc} %

\title{\LARGE \bf
IVRA: Improving Visual-Token Relations for Robot Action Policy\\
with Training-Free Hint-Based Guidance
}

\newif\ifanonymous
\anonymousfalse              

\ifanonymous
  \author{Anonymous Authors}
\else
  \author{ 
  {\bfseries
Jongwoo Park$^1$,
Kanchana Ranasinghe$^1$, 
Jinhyeok Jang$^2$,
    }\\[0.2em]
  {\bfseries
Cristina Mata$^1$,
Yoo Sung Jang$^1$,
Michael S Ryoo$^1$
    }\\[0.5em]
$^1$Stony Brook University \quad $^2$ETRI \vspace{0.2em} \\
\texttt{jongwopark@cs.stonybrook.edu} \\
}
\fi

\begin{document}
\maketitle
\thispagestyle{empty}

\begin{abstract}
Many Vision-Language-Action (VLA) models flatten image patches into a 1D token sequence, weakening the 2D spatial cues needed for precise manipulation. We introduce \textbf{IVRA}, a lightweight, training-free method that improves spatial understanding by exploiting affinity hints already available in the model's built-in vision encoder, without requiring any external encoder or retraining. IVRA selectively injects these affinity signals into a language-model layer in which instance-level features reside. This inference-time intervention realigns visual-token interactions and better preserves geometric structure while keeping all model parameters fixed. We demonstrate the generality of IVRA by applying it to diverse VLA architectures (\llara, OpenVLA, and FLOWER) across simulated benchmarks spanning both 2D and 3D manipulation (VIMA and LIBERO) and on various real-robot tasks. On 2D VIMA, IVRA improves average success by +4.2\% over the baseline \llara in a low-data regime. On 3D LIBERO, it yields consistent gains over the OpenVLA and FLOWER baselines, including improvements when baseline accuracy is near saturation (96.3\%\,$\rightarrow$\,97.1\%). Code and visualizations are available at: 
\href{https://jongwoopark7978.github.io/IVRA}{\texttt{jongwoopark7978.github.io/IVRA}}
\end{abstract}

\section{Introduction}
\label{sec:intro}
Vision-Language-Action (VLA) models have rapidly emerged as a promising approach for generating robot actions from images and natural-language instructions. Recent systems such as LLaRA~\citep{li2024llara}, OpenVLA~\citep{kim2024openvla}, FLOWER~\cite{reuss2025flower}, and LLARVA~\citep{niu2024llarva} pair large-scale pretrained vision encoders (e.g., CLIP or DINO~\citep{radford2021clip, caron2021dino}) with language models by flattening the 2D patch grid and appending the resulting visual tokens to the text sequence in a single Transformer pipeline. While this design leverages rich visual and linguistic knowledge, it also discards the image's native \emph{2D} neighborhood structure by treating visual patches as a 1D sequence of ``words.''

Flattening the patch grid into a 1D token sequence weakens local correlations and can blur object boundaries. This effect is visualized in \Cref{fig:overview}-(a): affinity maps computed from visual tokens inside the LLM are often diffuse and may bleed across object boundaries, whereas affinity maps from the vision encoder remain sharper and more instance-aligned. This suggests that instance-level features are diluted as flattened visual tokens repeatedly interact with text tokens through the LLM. Consequently, key cues such as object boundaries, attribute relations (e.g., color and shape), and fine-grained spatial relationships become harder to recover, hindering manipulation behaviors that require precise object interaction~\citep{pantazopoulos2024lost, pmlr-v164-shridhar22a, Ranasinghe2025PixelMA, han2025space}. Additional examples of blurred object boundaries in the baseline model’s LLM are shown in \Cref{fig:overview}-(b).

To counteract this loss of spatial context, existing solutions often resort to extensive retraining or specialized data. However, we propose a simpler and more lightweight method, \textbf{IVRA}, that augments VLA pipelines \emph{without} modifying their main components. Our approach uses an \emph{affinity map}, extracted from the model's encoder, indicating local similarity between patches. These \textbf{affinity hints} are then injected into deeper layers of the language model at inference time, re-weighting the flattened visual tokens based on their spatial correlations. This design restores fine-grained structure otherwise lost in the flattening step.

Experiments show that IVRA substantially improves instance-level recognition and localization across both simulated and real-world tasks. On VIMA-style manipulation benchmarks~\citep{jiang2023vima}, IVRA-equipped policies lead to higher success rates than methods that rely only on flattened embeddings. We further validate IVRA in real-world pick-and-place settings, where the ability to discriminate precise boundaries and attributes (e.g.\ color or shape) is critical for accurate grasping and placement.

Overall, this work contributes three key insights:
\emph{(1)}~it highlights how flattening vision tokens erodes 2D spatial cues in VLA architectures;
\emph{(2)}~it shows how the model's built-in encoder can serve as a source of affinity hints that recover local structure; and
\emph{(3)}~we show that injecting affinity hints into selected layers of the language model consistently improves diverse VLA models across multiple benchmarks and on both real and simulated robotic tasks, \emph{without} large-scale retraining or specialized data collection.

\section{Related Work}
\label{sec:related_work}
\subsection{Vision-Language-Action Models and Spatial Understanding}
Several works combine language models with visual backbones to handle open-domain perception and reasoning \citep{alayrac2022flamingo, li2023blip2, liu2023llava}. Many of these methods rely on globally pooled features from high-level layers in encoders such as CLIP \citep{radford2021clip}, which can lose spatial detail. Some studies suggest fusing earlier-layer or multi-level features to preserve finer object attributes \citep{caron2021dino, jiang2023clipdino}. Vision-language architectures extended to robot policies often require robust spatial reasoning to manipulate objects effectively. One example, VIMA, uses multimodal prompts for diverse manipulation tasks and benefits from strengthened instance-level visual understanding \citep{jiang2023vima}. 

Recently, open-vocabulary detection methods have been integrated into vision-language pipelines to improve spatial recognition \citep{Zhong_2022_CVPR, Guo_2024_CVPR,Ran2024LearningTL}. Several techniques focus on grounding language queries in specific regions, such as RegionCLIP and RegionGPT, which enhance or fine-tune CLIP-based backbones to detect arbitrary text-described objects \citep{Zhong_2022_CVPR, Guo_2024_CVPR}. These methods support fine-grained referencing across diverse vocabularies but typically require large-scale training on curated region annotations. Other approaches like Grounding~DINO incorporate language conditioning directly into the detection transformer to achieve open-set localization \citep{Liu_2024_ECCV}, also relying on extensive data for robust grounding.

\begin{figure*}[t]
    \centering

    \newlength{\overviewH}
    \setlength{\overviewH}{\heightof{\includegraphics[width=.58\textwidth]{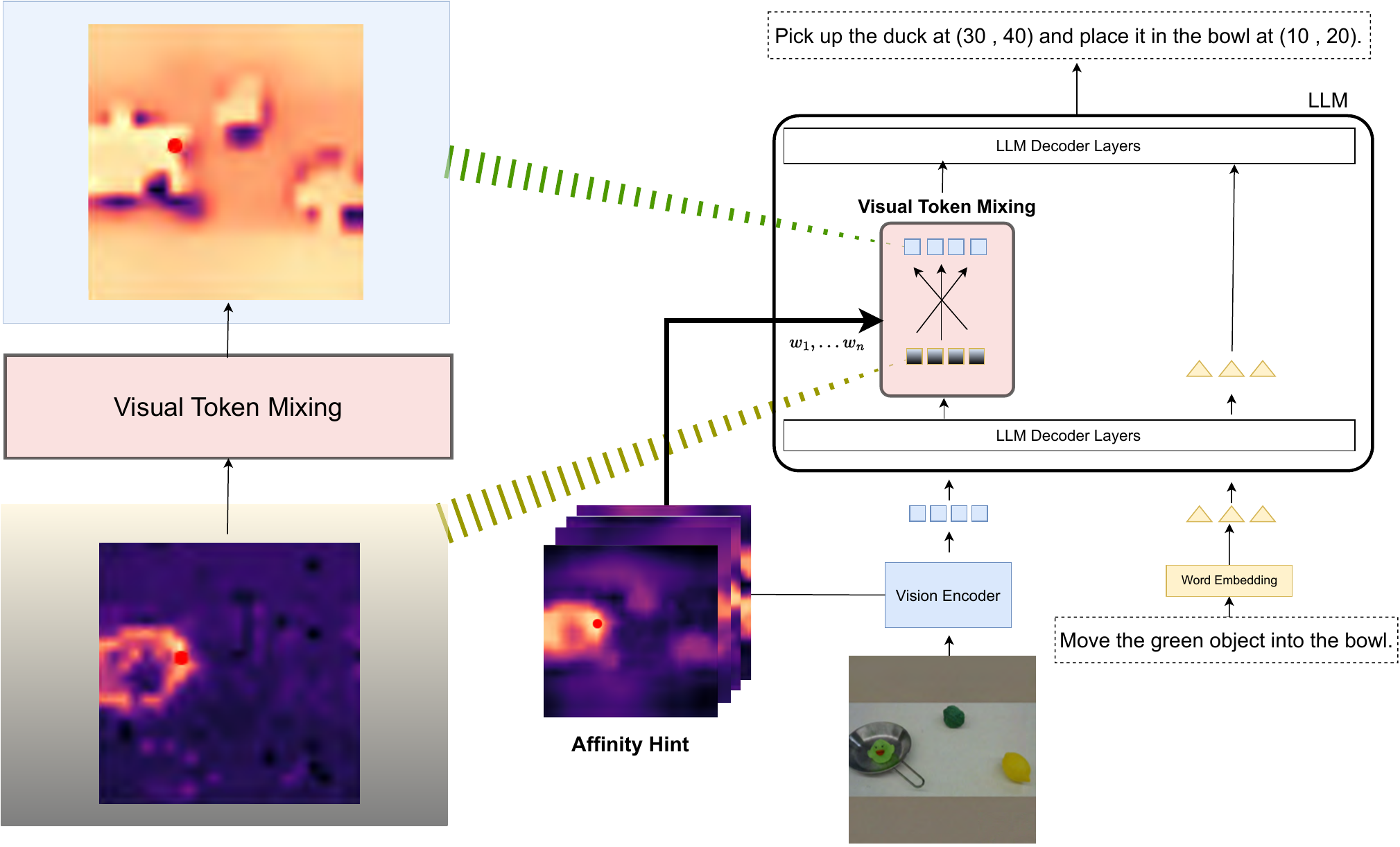}}}

    \begin{minipage}[c]{0.65\textwidth}
        \centering
        \includegraphics[width=\linewidth]{figures/overview_.pdf}
    \end{minipage}\hfill
    \begin{minipage}[c][\overviewH][c]{0.3\textwidth}
        \centering
        \includegraphics[width=\linewidth]{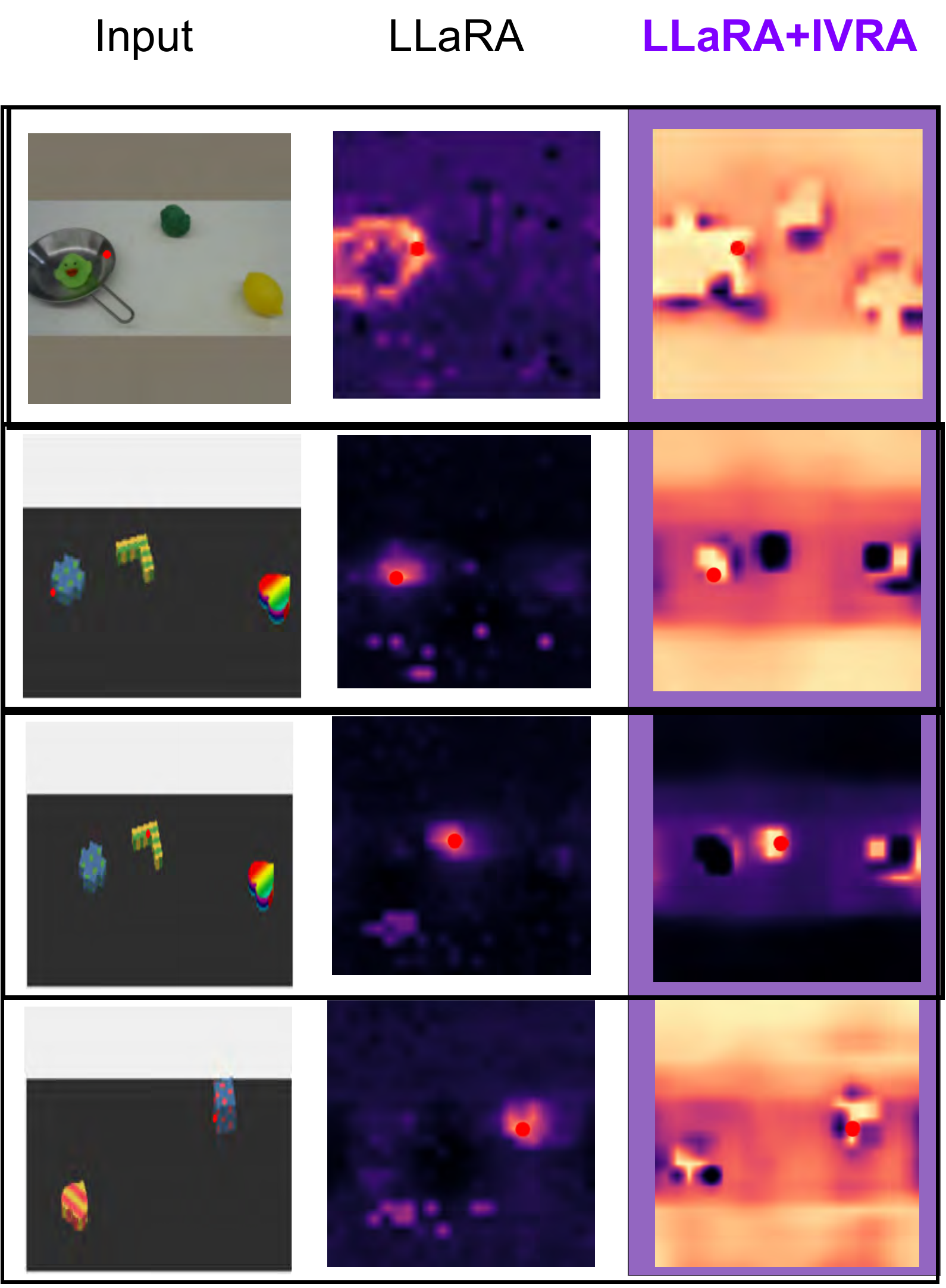}
    \end{minipage}

    \caption{
    \textbf{Main Overview of Our Method.} (a) \textbf{Left}: A frozen vision encoder provides an affinity hint that guides token mixing with weighted pooled tokens, preserving instance-level cues and improving manipulation policy quality. Brighter regions indicate higher affinity relative to the reference point (red dot). 
(b) \textbf{Right}: Affinity maps after applying IVRA shows sharper object boundaries and clearer object separation, aiding precise robot manipulation.
    }
    \label{fig:overview}
    \vspace{-0.5em}
\end{figure*}

\subsection{Affinity Hints and Instance-Level Feature Enhancement}
Several approaches inject affinity signals extracted from intermediate representations to enhance local structure in multimodal models \citep{zhou2024hints, wysoczanska2024clipdiy}. Such methods often compute patch-wise correlations and incorporate them into the token sequence, thereby emphasizing object boundaries and spatial configurations. Hints of Prompt applies a similar idea in driving scenarios, where hint tokens highlight object regions for more precise perception \citep{zhou2024hints}. ViP-LLaVA explores prompting methods that visually indicate key regions to large vision-language models, improving object reference resolution \citep{Cai_2024_CVPR}. Similarly, \citep{Xu_2022_CVPR, Liang_2023_CVPR, Ran2022PerceptualGI} learn to cluster or mask regions using language signals, enabling open-vocabulary segmentation with finer object detail. Though effective, these strategies often train adapters or decoders to fuse the spatial hints back into the model.

In the robotics domain, CLIPort uses CLIP’s global embeddings alongside a spatial “where” stream for manipulations, showing that bridging semantic and positional information fosters more generalized policies \citep{Shridhar_2022_CoRL}. PaLM-E further demonstrates large-scale multimodal pretraining can yield robust control policies once the model learns scene dynamics and object interactions \citep{Driess_2023_ICML}. However, these frameworks typically rely on extra task-specific training or specialized architecture modifications.

\subsection{Visual Affinity Maps in Multi-Modal Models}
Some lines of work leverage internal correlation maps from vision encoders to restore instance-level cues for downstream tasks \citep{jiang2023clipdino}. Techniques like CLIP-DIY produce denoised features by re-weighting coarse embeddings with affinity matrices, thus revealing finer-grained object representations \citep{wysoczanska2024clipdiy}. Unlike prior approaches, our method requires no external module to compute the affinity map and linearly mixes the original features with the affinity-pooled features, producing representations that preserve semantics while enriching object-level detail. RegionCLIP-like strategies facilitate region-level embeddings but rely on additional finetuning \citep{Zhong_2022_CVPR}. Grounding~DINO integrates text-based object queries directly into its detection modules \citep{Liu_2024_ECCV}, attaining strong open-world bounding box predictions through end-to-end training. While these approaches yield improved spatial understanding, they typically retrain entire pipelines on large-scale datasets. 

Our work shares the goal of preserving local structure but adopts a training-free approach, injecting affinity hints into mid-level LLM layers to rebalance object-specific features. This lightweight integration reclaims fine-grained information that standard global pooling discards, without extra training. By exploiting existing correlations in the vision encoder’s representations, IVRA retains strong semantic alignment from the original backbone and achieves better instance-level understanding in downstream action policies. This strategy balances spatial fidelity and high-level context, improving both recognition and manipulation tasks.

\section{Methodology}
\label{sec:method}
Our goal is to restore 2D spatial structure within Vision-Language-Action (VLA) models by injecting \emph{affinity hints} into selected layers of the LLM. The affinity hints are derived from the affinity maps extracted from the VLA's frozen vision encoder. We describe (i) how to extract the affinity map, (ii) how to apply affinity-guided pooling on visual tokens, and (iii) how we integrate this process into the VLA architecture. We illustrate this architecture in \Cref{fig:overview}-(a).

\subsection{Affinity Map Extraction}
\label{subsec:affinity-extraction}
Let an input image be divided into $N$ patches, and let $\{f_i\}_{i=1}^N$ denote the corresponding $d$-dimensional patch embeddings from a frozen vision encoder. To capture local relationships more effectively, we extract patch features from an intermediate layer (a few blocks before the final layer) of the vision encoder. Given any new image at inference time, we compute an affinity matrix $A \in \mathbb{R}^{N \times N}$ by
\begin{equation}
A_{ij} \;=\; \frac{f_i \cdot f_j}{\|f_i\| \,\|f_j\|},
\end{equation}
where $f_i \cdot f_j$ is the dot product between $f_i$ and $f_j$, and $\|\cdot\|$ denotes the Euclidean norm. A higher value of $A_{ij}$ indicates that patch $i$ and patch $j$ are likely to belong to the same object or share visual similarity. This affinity map serves as a patchwise connectivity prior, retaining the 2D spatial layout in a compact form.

\subsection{Affinity-Guided Visual Token Pooling}
\label{subsec:affinity-pooling}
Most VLA models flatten the $N$ patch embeddings into a 1D sequence of visual tokens and append them to text tokens in the LLM’s input stream. Let $v_i$ denote the feature for the $i$-th visual token at the start of layer $l$ of the LLM. Our approach injects the patchwise affinity information back into these visual tokens.

\vspace{2pt}
\noindent\textbf{Selecting Visual Tokens.}
In practice, a special token \texttt{<image>} indicates where the visual tokens begin. Once the actual patch embeddings replace \texttt{<image>}, they occupy indices in the middle of the token sequence. We keep track of these indices $\{i_1, \dots, i_N\}$ so we only update the visual tokens, leaving text tokens untouched.

\vspace{2pt}
\noindent\textbf{Weighted Average Pooling.}
Just before the layer’s self-attention block, we apply an affinity-guided pooling operation. For each visual token $v_i$, we compute a refined token $v_i'$ by mixing its neighbors according to:
\begin{equation}
v_i' \;=\; \sum_{j=1}^N \alpha_{ij}\, v_j, 
\quad \text{where} \quad
\alpha_{ij} \;=\; \frac{\max(A_{ij},\,0)}{\sum_{k=1}^N \max(A_{ik},\,0)}.
\end{equation}
Here, $A_{ij}$ is the affinity score between patches $i$ and $j$. Intuitively, patches that correlate strongly reinforce each other’s features, thereby preserving local spatial coherence. By re-weighting $v_i$ with contributions from visually similar patches, we restore some of the 2D structure that was lost in the flattening step. This operation modifies only the visual tokens; the textual tokens remain unchanged.

\subsection{Integration into VLA Models}
\label{subsec:integration}
We integrate the above pooling step into a few selected layers of the LLM, such as the 19th to 23rd layers. Concretely, these are the steps:

\begin{enumerate}
    \item \textbf{Vision Encoding:} An input image is split into $N$ patches and processed by the frozen vision backbone, yielding patch embeddings $\{f_i\}$.
    \item \textbf{Token Construction:} The patch embeddings are flattened into $\{v_i\}$ and inserted into the LLM token sequence at indices $\{i_1, \dots, i_N\}$, replacing the single \texttt{<image>} placeholder.
    \item \textbf{Affinity-Guided Pooling:} Before the self-attention sublayer at each chosen layer $l$, we update each $v_i$ via the weighted average:
    \[
    v_i' = \sum_{j=1}^N \alpha_{ij}\,v_j,\quad 
    \alpha_{ij} = \frac{\max(A_{ij},\,0)}{\sum_{k=1}^N \max(A_{ik},\,0)}.
    \]
    \item \textbf{Token Mixing:} To preserve the semantics of the original token while injecting object-aware evidence from pooling, we linearly blend the pooled token with its unpooled counterpart. We form the final visual token as a convex combination:
    \[
        v_i^{\mathrm{mix}} = (1-\lambda)\, v_i + \lambda\, v_i',\qquad \lambda \in [0,1].
    \]
    \item \textbf{Continuing the LLM:} The updated visual tokens $v_i^{\mathrm{mix}}$ proceed through the layer normalization, self-attention, and subsequent transformations as usual. Text tokens are not modified.
    \item \textbf{Output Decoding:} Finally, the LLM produces the next-stage representations for generating policy actions or textual responses, depending on the specific VLA setup.
\end{enumerate}

\begin{table*}[t]
\centering
\small
  \label{tab:vima_short}
  \centering
  \caption{
\textbf{VIMA-Bench Results:}
Average success rate (\%) on the four VIMA generalization tasks. \textbf{LLaRA+IVRA} consistently improves over the \llara baseline and outperforms both \llara and VIMA across all tasks. It indicates IVRA's robust instance-level generalization. As reported in \llara, we \textcolor{down}{downplayed VIMA's performance on the Novel Task} due to missing data in the original VIMA dataset.}
  \scalebox{0.83}{ 
  \begin{tabular}{l|c|c|lllll}
    \toprule
    \textbf{Method} & \textbf{Ext. Module} & \textbf{Data} & \textbf{Novel Task(\%)} &  \textbf{Novel Object(\%)} &  \textbf{Obj. Comb.(\%)} & \textbf{Obj. Place.(\%)} & \textbf{Avg.(\%)}\\
    \midrule
    \llara~\cite{li2024llara} & \multirow{2}{*}{\cellcolor{white}N/A} &   12\% & 22.5 & 57.1 &  66.5 &  69.6 & 53.9 \\
    \cellcolor{myblue}\textbf{\modelname} (Ours) &  &   
    \cellcolor{myblue}12\% & 
    \cellcolor{myblue}\textbf{27.5\inc{5.0}} & 
    \cellcolor{myblue}\textbf{61.3\inc{4.2}} &  
    \cellcolor{myblue}\textbf{70.4\inc{3.9}} &  
    \cellcolor{myblue}\textbf{73.1\inc{3.5}} &
    \cellcolor{myblue}\textbf{58.1}\inc{4.2} \\
    
    \midrule
    VIMA~\citep{jiang2023vima} & \multirow{3}{*}{\cellcolor{white}Oracle Det.} & 100\%  &  \textcolor{down}{48.8} & 77.9 & 81.9 & 80.7 & 72.3 \\
    \llara~\cite{li2024llara}  &  & 12\% & 33.8 & 79.2 & 88.1 &  90.0 & 72.8 \\
    \cellcolor{myblue}\textbf{\modelname} (Ours) &  & 
    \cellcolor{myblue}12\%  & 
    \cellcolor{myblue}\textbf{37.5\inc{3.7}} & 
    \cellcolor{myblue}\textbf{80.8\inc{1.6}} &
    \cellcolor{myblue}\textbf{88.5}\inc{0.4}  &  
    \cellcolor{myblue}\textbf{90.4\inc{0.4}}  &
    \cellcolor{myblue}\textbf{74.3}\inc{1.5} \\
    \bottomrule
\end{tabular}
}
\vspace{0.5em}
\label{table:vima}
\end{table*}

\section{Experimental Results}
\subsection{VIMA Simulated Environment}
We incorporated the affinity hint into \textbf{\llara}~\cite{li2024llara} and evaluated it on the four VIMA~\citep{jiang2023vima} tasks: Novel Task, Novel Object, Object Combination (novel combination of seen objects and textures), and Object Place (placement generalization with seen objects). We ordered these partitions by increasing ease of generalization, with Novel Task being the most challenging. 

VIMA is trained on 660k expert trajectories in total across 17 task templates. \llara uses 80k trajectories (approximately 12\% of total trajectories) and evaluate each task using 20 randomized seeds, reporting mean success. As reported in \llara, we de-emphasize VIMA’s Novel Task due to missing data in the original dataset. 

From \Cref{table:vima}, \modelname surpasses \llara in both without and with oracle detector. Without the oracle detector, \modelname improves over \llara by +5.0\% (Novel Task), +4.2\% (Novel Object), +3.9\% (Object Combination), and +3.5\% (Object Place). With the oracle detector, the gains remain positive across all four tasks: +3.7\%, +1.6\%, +0.4\%, and +0.4\%, respectively. Notably, using only 12\% of the data, our oracle-detector variant also exceeds VIMA (trained with 100\% data) across all four tasks, underscoring the robustness and generalizability.

\subsection{LIBERO Simulated Environment}
To assess whether IVRA extends beyond 2D image-based manipulation (VIMA) to \emph{3D} embodied control, we evaluate on the LIBERO benchmark. LIBERO is a suite of language-conditioned 3D manipulation tasks that stresses different forms of transfer and compositionality. The suites isolate complementary skills: \emph{Spatial} emphasizes spatial relations, \emph{Object} emphasizes object-centric manipulation, \emph{Goal} varies target goals, and \emph{Long} requires multi-step temporal composition. By following \textbf{OpenVLA}~\cite{kim2024openvla} evaluation, we report results on the standard task suites: \emph{Goal}, \emph{Object}, \emph{Spatial}, as well as \emph{Long}. In addition, we also report \emph{LIBERO-90}, a large collection of 90 short-horizon tasks from the LIBERO-100 suite, when comparing against \textbf{FLOWER}~\cite{reuss2025flower}

We insert IVRA into OpenVLA \emph{at inference time only}, without retraining or modifying the base model. As summarized in \Cref{table:libero-res}, IVRA consistently boosts OpenVLA across all LIBERO suites, increasing the overall average from 76.5\% to 77.6\% (+1.1\%). This confirms that injecting affinity hints helps not only in 2D settings but also for 3D spatial reasoning and long-horizon control. Beyond improving OpenVLA itself, OpenVLA+IVRA also surpasses other baselines evaluated under the same protocol, including Diffusion Policy~\cite{chi2023diffusion} and Octo~\cite{octo_2023}, by +5.2\% and +2.5\% in average success, respectively.

To further probe architectural generality, we apply IVRA to \textbf{FLOWER} in a plug-and-play manner (again, no additional training). For a fair comparison, we keep hyperparameters identical to each baseline and only set the token-mixing coefficient \(\lambda\) to 0.2 for FLOWER+IVRA. Despite FLOWER's \emph{very strong} baselines 94\%-99\% across LIBERO categories—IVRA still yields consistent gains in every setting, e.g., Task-\textit{90}: 93.4\%\,$\rightarrow$\,96.0\% (+2.6\%), Task-\textit{Object}: 99.3\%\,$\rightarrow$\,99.9\% (+0.6\%) and an overall lift from 96.3\% to 97.1\% (+0.8\%). Improvements at such near-saturated accuracies indicate that IVRA contributes complementary structure rather than merely compensating for weak baselines.

The results of 2D VIMA and 3D LIBERO highlight IVRA's broad generalization: (i) across input dimensionality (2D and 3D tasks), (ii) across VLA architectures (OpenVLA and FLOWER), and (iii) across baseline accuracy regimes—from challenging mid-50\% ranges to high-90\% near-saturation. Importantly, all gains are achieved via a lightweight inference-time modification with no retraining.

\begin{table}[t]
  \centering
  \caption{
  \textbf{LIBERO Benchmark Results}: Average success (\%) on LIBERO tasks. IVRA yields consistent improvements for both OpenVLA and FLOWER, even when baseline accuracy is near saturation
  }
  \label{table:libero-res}
  \setlength{\tabcolsep}{3pt}
  \renewcommand{\arraystretch}{1.05}
  \resizebox{\columnwidth}{!}{%
    \begin{tabular}{lllllll}
      \toprule
      Method & 90 & Goal & Object & Long & Spatial & Average \\
      \midrule
      Diffusion Policy~\cite{chi2023diffusion}  & - & 68.3 & 92.5 & 50.5 & 78.3 & 72.4 \\
      Octo~\cite{octo_2023}              & - & 84.6 & 85.7 & 51.1 & 78.9 & 75.1 \\
      \midrule
      OpenVLA~\cite{kim2024openvla}          & -  & 79.2 & 88.4 & 53.7 & 84.7 & 76.5 \\
      \rowcolor{myblue}
      \textbf{OpenVLA+IVRA} (Ours)  & - & \textbf{81.2}\inc{2.0} & \textbf{89.6}\inc{1.2} & \textbf{54.2}\inc{0.5} & \textbf{85.5}\inc{0.8} & \textbf{77.6}\inc{1.1} \\    
      \midrule
      FLOWER~\cite{reuss2025flower} & 93.4 & 96.9 & 99.3 & 94.5 & 97.2 & 96.3 \\
      \rowcolor{myblue}
      \textbf{FLOWER+IVRA} (Ours)      & \textbf{96.0}\inc{2.6} & \textbf{97.6}\inc{0.7} & \textbf{99.9}\inc{0.6} &
                                 \textbf{94.9}\inc{0.4} & \textbf{97.3}\inc{0.1} & \textbf{97.1}\inc{0.8} \\
      \bottomrule
    \end{tabular}%
  }
\end{table}

\subsection{Real World Environment}
\begin{figure}[t]
  \centering
  \includegraphics[width=0.7\columnwidth]{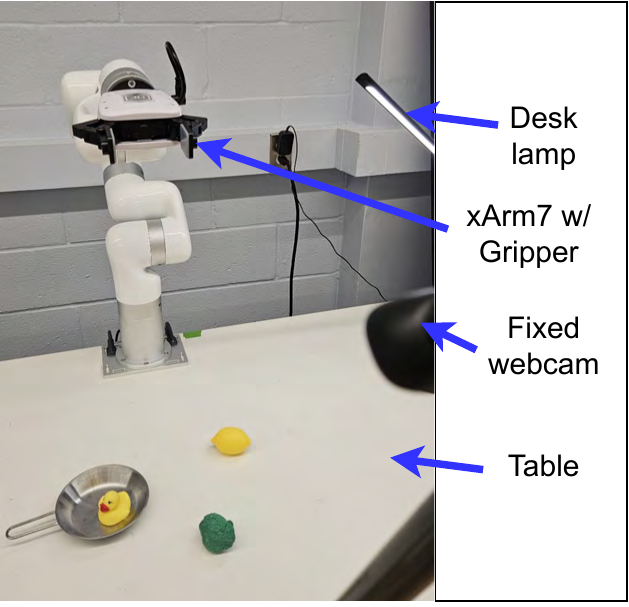}
  \caption{Gripper-equipped arm and overhead RGB camera.}
  \label{fig:xarm}
\end{figure}

We further conduct real-world experiments involving zero-shot generalization on a novel robotic setup. Similar to the setting in LLaRA~\citep{li2024llara}, our environment consists of a robot arm with a gripper, positioned under a fixed RGB camera for collecting observations (see ~\Cref{fig:xarm}). As the objects are placed on a plain surface and the camera is stationary, a simple linear mapping between the image coordinates and the robot action space can be established by calibration. We use a policy (inBC-8k model) that was trained purely on synthetic data on four real-world tasks, T1-T4, defined as follows:
\begin{itemize}
    \item \textbf{Target Object (T1):} ``Pick up \{\texttt{object}\} and drop it into a pan." Here, \{\texttt{object}\} is chosen from nine toy items (\emph{duck, corn, pepper, lemon, eggplant, orange, potato, broccoli, strawberry}). Multiple objects are sparsely placed on the table. This tests the model's capacity to choose the correct object (i.e., picking and placing a specified object). See \Cref{fig:tobjvis} for visualization.
    
    \item \textbf{Color Match (T2):} ``Pick up the object the same color as \{\texttt{object\_ref}\} and drop it into a pan." Here, \{\texttt{object\_ref}\} has three possible colors (\emph{yellow, orange, green}), prompting the robot to identify an item \emph{by its color} and then place it inside the pan. This tests instance-level attribute understanding and more semantic scene comprehension. See \Cref{fig:cmvis} for visualization.
    
    \item \textbf{Cluttered Localization (T3):} ``Pick up the \{\texttt{object}\} and drop it into a pan.'' The object set and prompt follow \textbf{T1}, but the target object is initialized very close to other objects (distractors). Neighboring items are randomly placed around the target, both horizontally (left/right) and vertically (top/bottom). This tests the model's ability to localize the target object under clutter conditions. See \Cref{fig:locvis} for visualization.
    
    \item \textbf{Relative Height (T4):} ``Pick up the short (or long) object and place it on the pan.'' Here, multiple objects of varying lengths from \textbf{T1} are randomly placed in the scene (e.g., one short and two long, or vice versa). This tests comparative size/height understanding. See \Cref{fig:ohvis} for visualization.
\end{itemize}

\begin{table}[t]
  \centering
  \caption{\textbf{Real-world Results:} LLaRA+IVRA's zero-shot performance across four tasks. IVRA substantially improves the baseline LLaRA across all tasks T1 (Target Object), T2 (Color Match), T3 (Cluttered Localization), T4 (Relative Height).} %
  \label{table:real}                                            %
  \def\arraystretch{1.3}
  \setlength\tabcolsep{0.5em}
  \resizebox{\columnwidth}{!}{%
      \begin{tabular}{l|l|llll}
        \toprule
        \textbf{Model} & \textbf{Trained Type} & \textbf{T1} & \textbf{T2} & \textbf{T3} & \textbf{T4} \\
        \midrule
        RT\textendash 2\textendash Style & VIMA\textendash 8k   & 0  & 0  & 0  & 0  \\
        RT\textendash 2\textendash Style & VIMA\textendash 660k & 0  & 0  & 0  & 0  \\
        \midrule
        LLaRA~\cite{li2024llara}            & InBC\textendash 8k & 50 & 30 & 45 & 50 \\
        \rowcolor{myblue}
        \textbf{LLaRA+IVRA} (Ours) & InBC\textendash 8k & \textbf{60}\inc{10} & \textbf{60}\inc{30} & \textbf{75}\inc{30} & \textbf{70}\inc{20} \\
        \bottomrule
      \end{tabular}
    }
\end{table}

\begin{figure*}[t]
\centering
    \includegraphics[width=0.8\linewidth]{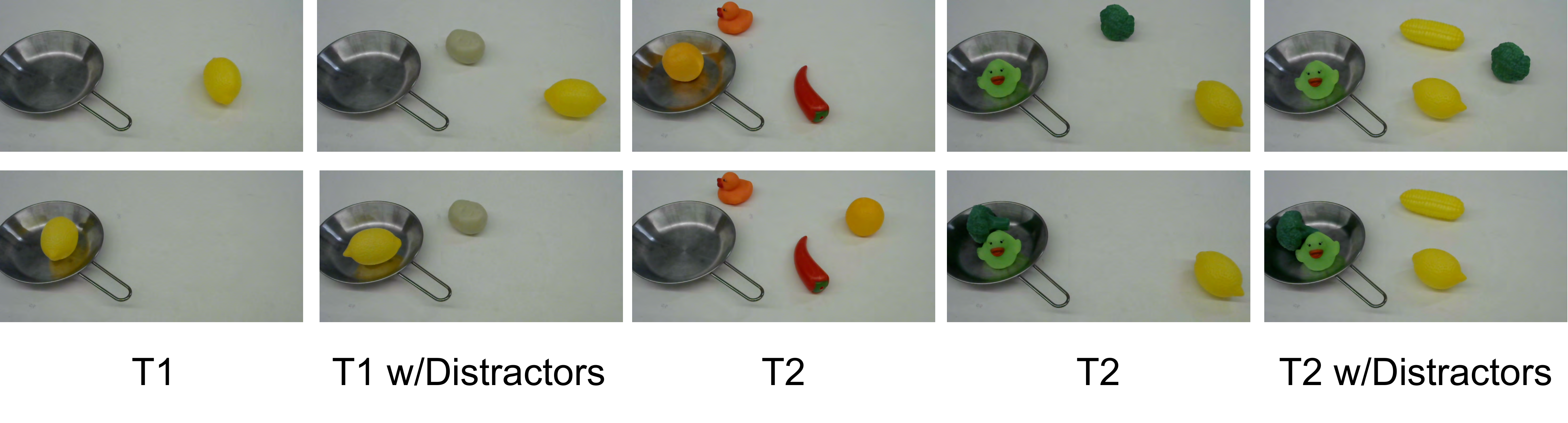}
    \caption{Visualization of Real-world Scenarios: The top row shows the initial scene and the bottom row the corresponding successful end state for each of the three tasks. Additional columns include trials with distracting objects in the workspace.
    }
    \vspace{-0.5em}
    \label{fig:realstfi}
\end{figure*}

All objects in each episode are placed randomly on a tabletop, and none of these real-world toys appear in the training set. A success in T1 is determined by whether the specified object is placed fully inside the pan. A success in T2 requires the robot to correctly identify an object of the same color as the reference toy and place it in the pan. T3 fails if the robot touches a neighboring object. Across 10 episodes, each episode is initialized randomly, yet within a given episode the task setup is identical for all models. We allow at most 5 retries per episode; a retry occurs when the model fails to produce an action given the input (note VIMA allows 7 retries per episode but we shorten in the interest of time). We report the average success rate over 10 episodes. We present representative scenarios from our real-world experiments in \Cref{fig:realstfi}: T1, T1 with obstacles, T2, and T2 with obstacles.

\paragraph{Zero-shot Generalization.}
To evaluate zero-shot generalization, we use the LLaRA model pretrained on only 1.2\% of the VIMA training data (inBC-8k), thereby creating a more challenging environment for adaptation. We also include RT-2-Style, a LLaRA variant, as shown in Table~\ref{table:real}. RT-2 Style is the modification of RT-2 training recipe~\citep{brohan2023rt} in such a way that VLM produces a discrete set of special tokens directly mappable to quantized robot actions, instead of continuous outputs. T1 (placing a specific object in the pan) is relatively simpler, akin to the ``Object Place" task in VIMA. Meanwhile, T2, T3, T4 is more challenging, similar in spirit to the ``Novel Task" scenario in VIMA. 

In Table~\ref{table:real}, we observe that \textbf{LLaRA+IVRA} outperforms vanilla LLaRA in all four tasks T1-T4 achieving a +10\% improvement for the simpler T1 and a notable up to +30\% boost for the more challenging T2-T4. This trend aligns with the findings from~\Cref{table:vima}, where IVRA consistently yields greater performance gains in more difficult tasks (\emph{e.g.}, +5.0\% on Novel Task vs. +3.5\% on Object Place). In T2, IVRA's instance-level affinity hint appears crucial for recognizing and localizing an object of the correct color, thereby enhancing its zero-shot reasoning ability. For T3 \textbf{LLaRA+IVRA} obtains +30\% gains underscores the benefits of IVRA for precise object localization and accurate task execution. For T4 \textbf{LLaRA+IVRA} achieves 70\% accuracy versus 50\% with LLaRA, highlighting a 20\% improvement. This result further demonstrates the instance-level grounding capability and generalization benefits conferred by IVRA.

\subsection{Qualitative Results}
\label{sec:qual_results}
\Cref{fig:overview}-(b) shows qualitative comparisons on both real-world and simulated tasks, illustrating how \emph{IVRA} refines object-level recognition within a \emph{visual-language-action} model. Each column means (1) the input image, (2) affinity maps \emph{before} IVRA, and (3) affinity maps \emph{after} IVRA. Affinity map is drawn relative to the reference point (red dot) and brighter areas denote stronger affinity with that reference point.

\paragraph{Observations}
Affinity maps of the visual tokens in the baseline models are often noisy or incomplete, making it difficult to delineate object boundaries—particularly when objects share similar colors or when the reference point lies on an edge. Once IVRA is applied, the affinity maps become much more coherent, clearly highlighting individual objects rather than scattering activations throughout the background. This finer-grained delineation is consistent across both real and simulated environments, aligning with the denoising effects observed in related work~\citep{wysoczanska2024clipdinoiser}.

\paragraph{Impact on Performance}
These qualitative findings corroborate IVRA’s strong quantitative gains in both real-world and simulated scenarios (\emph{cf.}~\Cref{table:vima,table:libero-res,table:real}). By supplying an instance-level affinity hint, IVRA helps the model better understand the similarity and differences between objects, which is crucial for tasks demanding precise physical interactions—like grasping and placement. In real-world T2, for example, locating and picking an item of the correct color benefits significantly from sharper boundary recognition, contributing to a +30\% increase over LLaRA. Similarly, in the VIMA tasks, the largest improvements appear on the more difficult Novel Task, where object-specific token features are most essential. Taken together, these results illustrate how IVRA bolsters robust object localization and manipulation within a visual-language-action framework, ultimately translating into high success rates in real robot experiments and complex simulated benchmarks.

\section{Visualization of Real World Experiments}
\begin{figure*}[!t]
    \centering
    \captionsetup[subfigure]{labelformat=empty}

    \begin{subfigure}[t]{0.49\textwidth}
        \centering
        \begin{tabular}{@{}c@{}c@{}c@{}}
            \includegraphics[width=0.33\textwidth]{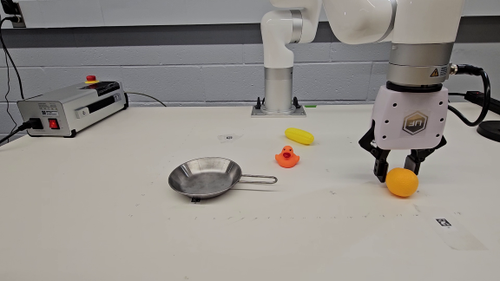} &
            \includegraphics[width=0.33\textwidth]{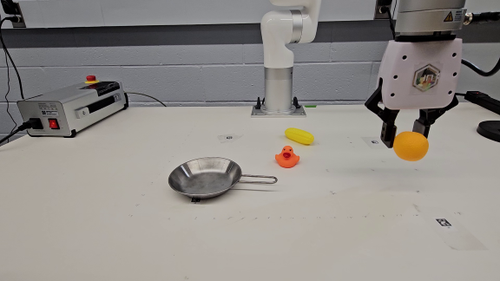} &
            \includegraphics[width=0.33\textwidth]{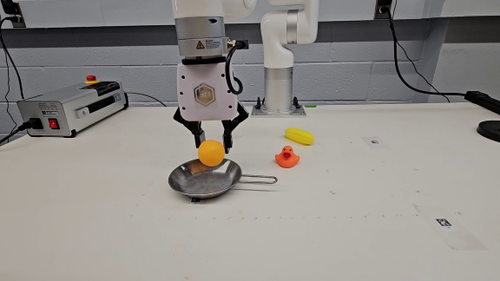} \\
            \includegraphics[width=0.33\textwidth]{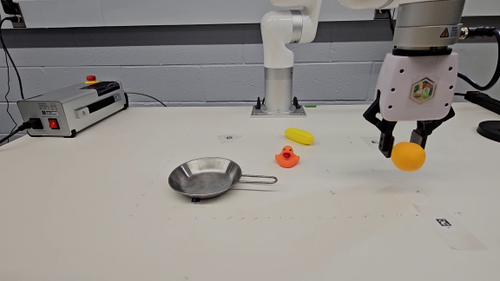} &
            \includegraphics[width=0.33\textwidth]{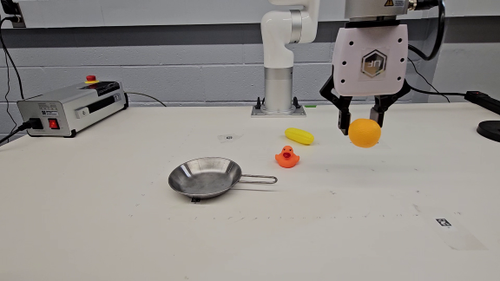} &
            \includegraphics[width=0.33\textwidth]{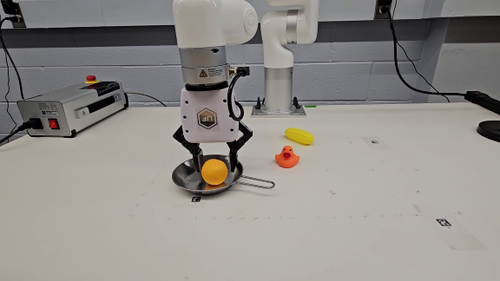} \\
        \end{tabular}
        \caption{\textbf{Target Object (T1).} ``Pick up the orange and
        drop it into a pan.''}
        \label{fig:tobjvis}
    \end{subfigure}
    \hfill
    \begin{subfigure}[t]{0.49\textwidth}
        \centering
        \begin{tabular}{@{}c@{}c@{}c@{}}
            \includegraphics[width=0.33\textwidth]{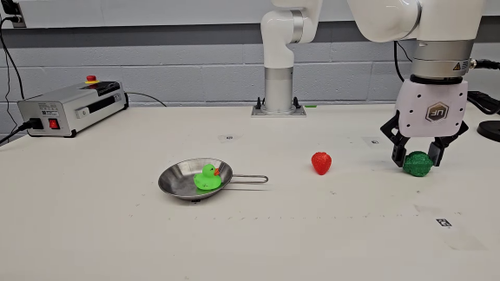} &
            \includegraphics[width=0.33\textwidth]{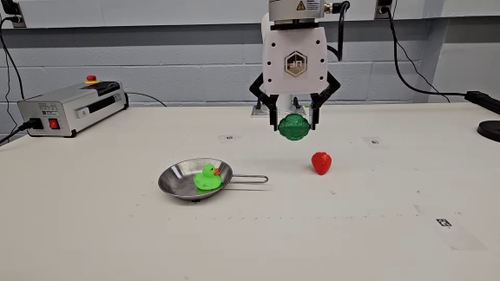} &
            \includegraphics[width=0.33\textwidth]{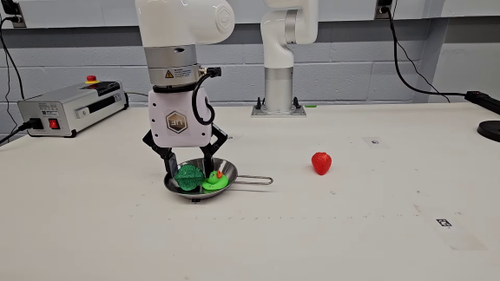} \\
            \includegraphics[width=0.33\textwidth]{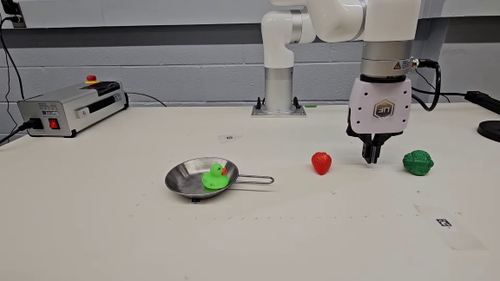} &
            \includegraphics[width=0.33\textwidth]{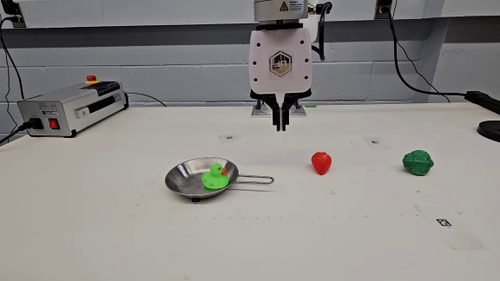} &
            \includegraphics[width=0.33\textwidth]{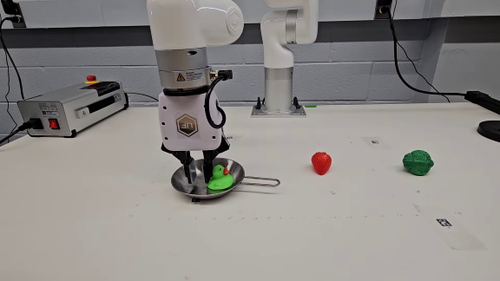} \\
        \end{tabular}
        \caption{\textbf{Color Match (T2).} ``Pick up object same color
        as the duck and drop it into a pan.''}
        \label{fig:cmvis}
    \end{subfigure}

\end{figure*}

\begin{figure*}[!t]
    \centering
    \captionsetup[subfigure]{labelformat=empty}
    
    \begin{subfigure}[t]{0.49\textwidth}
        \centering
        \begin{tabular}{@{}c@{}c@{}c@{}}
            \includegraphics[width=0.33\textwidth]{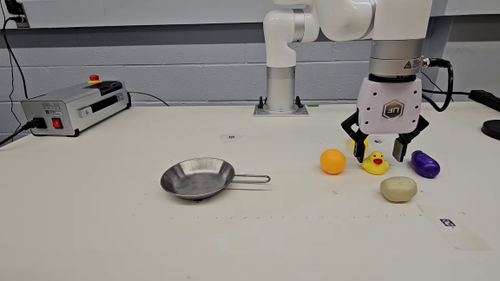} &
            \includegraphics[width=0.33\textwidth]{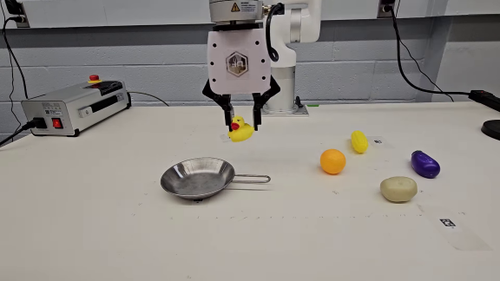} &
            \includegraphics[width=0.33\textwidth]{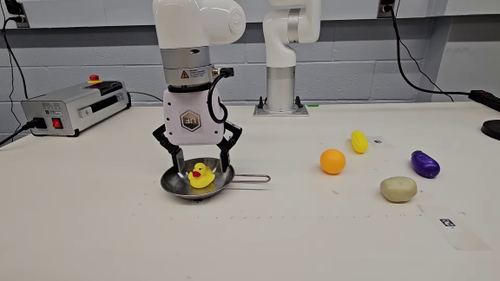} \\
            \includegraphics[width=0.33\textwidth]{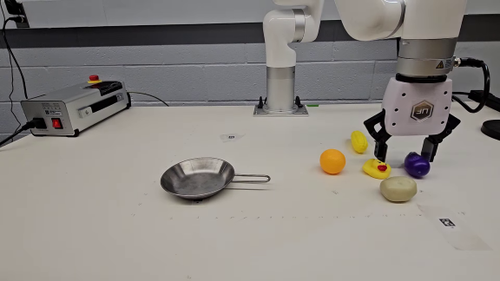} &
            \includegraphics[width=0.33\textwidth]{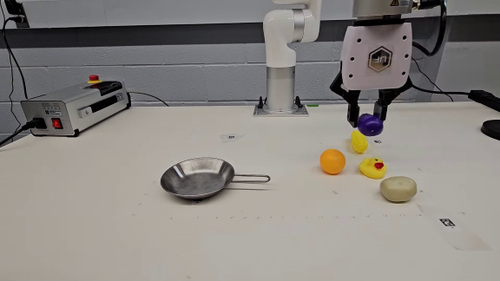} &
            \includegraphics[width=0.33\textwidth]{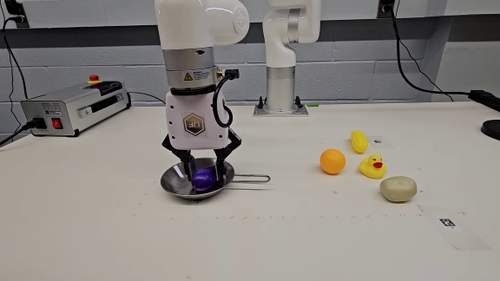} \\
        \end{tabular}
        \caption{\textbf{Localization (T3).} '`Pick up the yellow duck and drop it into a pan.''}
        \label{fig:locvis}
    \end{subfigure}
    \hfill
    \begin{subfigure}[t]{0.49\textwidth}
        \centering
        \begin{tabular}{@{}c@{}c@{}c@{}}
            \includegraphics[width=0.33\textwidth]{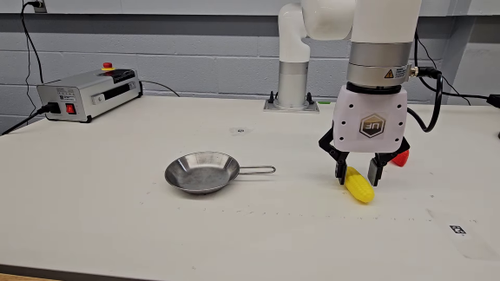} &
            \includegraphics[width=0.33\textwidth]{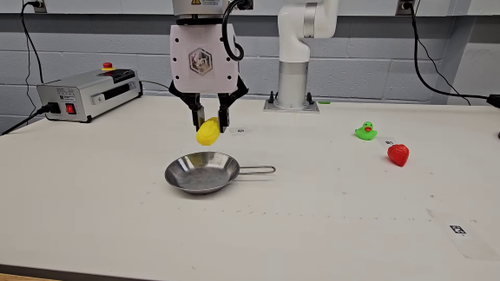} &
            \includegraphics[width=0.33\textwidth]{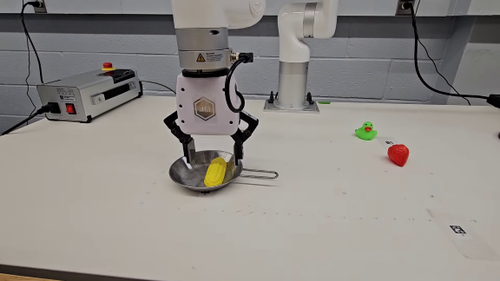} \\
            \includegraphics[width=0.33\textwidth]{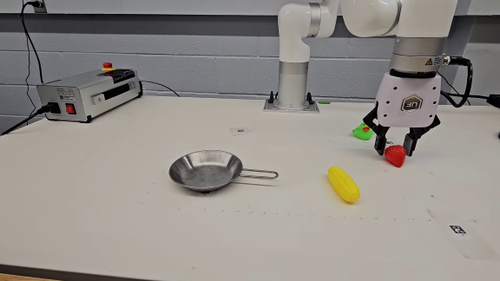} &
            \includegraphics[width=0.33\textwidth]{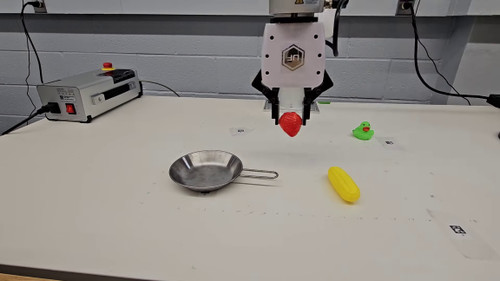} &
            \includegraphics[width=0.33\textwidth]{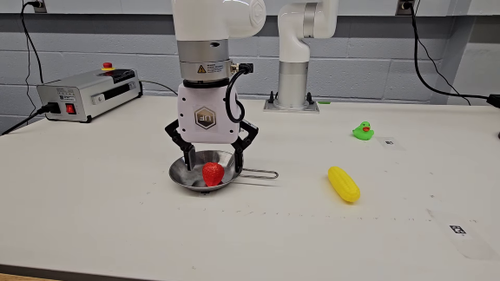} \\
        \end{tabular}
        \caption{\textbf{Object Height (T4).} ``Pick up the long object and place it on the pan.''}
        \label{fig:ohvis}
    \end{subfigure}

\end{figure*}

Figures~\ref{fig:tobjvis},~\ref{fig:cmvis},~\ref{fig:locvis}, and \ref{fig:ohvis} show the robot trajectories for one moderate task (T1) and three challenging tasks (T2-T4). In every figure, the top row depicts the behavior under LLaRA+IVRA, while the bottom row illustrates the performance of the original LLaRA approach. In Figure~\ref{fig:tobjvis} showing T1, both LLaRA and LLaRA+IVRA pick up the correct object as this task is relatively moderate. Figure~\ref{fig:cmvis} demonstrates the T2: LLaRA+IVRA correctly identifies and picks up the green broccoli (matching the duck’s color), whereas LLaRA fails to pick up the broccoli. Figure~\ref{fig:locvis} corresponds to localization task T3, showing LLaRA+IVRA successfully grasps the specified target (the yellow duck), whereas LLaRA incorrectly selects the eggplant instead. In Figure~\ref{fig:ohvis} showing T4, LLaRA+IVRA correctly picks up the corn (the long object), while LLaRA mistakenly chooses a strawberry (the short object).

Overall, these four real-world tasks in addition to the simulation results show that IVRA provides stronger instance-level understanding, allowing more accurate localization and attribute-based object selection in novel, real-world contexts. The significant performance gains on \emph{Localization}, \emph{Object Height}, and \emph{Object Color Matching} tasks suggest that IVRA helps the model to better interpret and resolve more intricate scenes and instructions.

\section{Complexity Analysis \& Ablations}
\begin{table}[h]
  \centering
  \caption{Runtime and parameter overhead of IVRA. IVRA introduces only a marginal runtime overhead and does not introduce any additional parameters.}
  \label{table:complexity}
  \setlength{\tabcolsep}{6pt}        %
  \renewcommand{\arraystretch}{1.1}  %
  \resizebox{0.7\columnwidth}{!}{%
    \begin{tabular}{lcc}
      \toprule
      Model & Latency (s) & Param. (GB) \\
      \midrule
      LLaRA            & 2.00 & 15.8 \\
      \rowcolor{myblue}
      \textbf{LLaRA\,+\,IVRA}   & 2.06 & 15.8 \\
      \bottomrule
    \end{tabular}
  }
\end{table}

\subsection{Runtime Analysis}
\Cref{table:complexity} shows the complexity analysis of the IVRA. We tested it on LLaRA and with one NVIDIA RTX 6000 Ada GPU. IVRA only adds small latency (3\%) and do not introduce any parameters as it does not use external modules.

\subsection{Ablation Study}
\begin{table}[t]
\centering
\caption{\textbf{Ablation on Affinity-based Pooling in IVRA.} We explore different (a) single-layer placements, (b) numbers of consecutive layers, (c) within-layer locations for our affinity-based weighted average pooling when applied to LLaRA, (d) token-mixing coefficient \(\lambda\). Tasks are \textbf{NT} = Novel Task, \textbf{NO} = Novel Object, \textbf{OC} = Object Combination, and \textbf{OP} = Object Placement from VIMA~benchmark. The row in blue is our final choice.}
\footnotesize
\setlength{\tabcolsep}{3pt}

\begin{minipage}{0.48\linewidth}
\centering
\scalebox{0.90}{
\begin{tabular}{c|cccc}
\toprule
\textbf{Layer Pos.} & \textbf{NT} & \textbf{NO} & \textbf{OC} & \textbf{OP} \\
\midrule
31  & 22.5 & 58.3 & 67.7 & 69.6 \\
23  & 23.8 & 57.9 & 66.9 & 71.2 \\
22  & 26.2 & 59.2 & 68.1 & 71.9 \\
21  & 25.0 & 60.0 & 67.7 & 72.3 \\
\rowcolor{myblue}
20  & 27.5 & 61.3 & 70.4 & 73.1 \\
19  & 22.5 & 60.8 & 68.1 & 72.3 \\
11  &  22.5 & 58.3 &  64.2 &  71.9 \\
A. Proj. & 0.0 & 5.8 & 5.0 & 5.8 \\
B. Proj. & 0.0 & 5.4 & 3.8 & 6.9 \\
\bottomrule
\end{tabular}}
\vspace{0.3em}
\subcaption{Single-layer Performance}
\label{tab:subtable1}
\end{minipage}
\hfill
\begin{minipage}{0.48\linewidth}
\centering
\scalebox{0.90}{
\begin{tabular}{c|cccc}
\toprule
\textbf{\# Layers} & \textbf{NT} & \textbf{NO} & \textbf{OC} & \textbf{OP} \\
\midrule
\rowcolor{myblue}
1  & 27.5 & 61.3 & 70.4 & 73.1  \\
2  & 26.2 & 60.8 & 69.6 & 73.8 \\
3  & 27.5 & 59.6 & 68.5 & 72.3 \\
4  & 27.5 & 59.2 & 68.8 & 72.3 \\
5  & 27.5 & 59.2 & 69.2 & 72.7 \\
\bottomrule
\end{tabular}}
\vspace{0.3em}
\subcaption{Multi-layer Performance}
\label{tab:subtable2}
\end{minipage}

\vspace{0.9em}
\begin{minipage}{0.48\linewidth} %
\centering
\scalebox{0.90}{
\begin{tabular}{c|cccc}
\toprule
\textbf{Pool. Loc.} & \textbf{NT} & \textbf{NO} & \textbf{OC} & \textbf{OP} \\
\midrule
\rowcolor{myblue}
P0 & 27.5 & 61.3 & 70.4 & 73.1  \\
P1 & 22.5 & 56.7 & 68.1 & 70.8 \\
P2 & 22.5 & 56.7 & 66.5 & 70.8 \\
P3 & 25.0 & 60.8 & 68.1 & 72.7 \\
P4 & 23.8 & 57.1 & 66.5 & 70.4 \\
\bottomrule
\end{tabular}}
\vspace{0.3em}
\subcaption{W.Avg Pooling Locations}
\label{tab:subtable3}
\end{minipage}
\hfill
\vspace{0.9em}
\begin{minipage}{0.48\linewidth}
\centering
\scalebox{0.90}{
\begin{tabular}{c|cccc}
\toprule
\textbf{$\lambda$} & \textbf{NT} & \textbf{NO} & \textbf{OC} & \textbf{OP} \\
\midrule
1   & 30   & 58.3 & 70.4 & 71.2 \\
0.7 & 30   & 58.8 & 68.8 & 74.2 \\
\rowcolor{myblue}
0.3 & 27.5 & 61.3 & 70.4 & 73.1 \\
0   & 22.5 & 57.1 & 66.5 & 69.6 \\
\bottomrule
\end{tabular}}
\vspace{0.3em}
\subcaption{Token-Mixing Coefficient $\lambda$}
\label{tab:subtable4}
\end{minipage}

\label{table:three-subtables}
\end{table}

Table~\ref{table:three-subtables} summarizes our ablation results on the VIMA benchmark with an oracle detector. We report success rates (\%) for four tasks: \textbf{NT} (Novel Task), \textbf{NO} (Novel Object), \textbf{OC} (Object Combination), and \textbf{OP} (Object Placement). In all sub-tables, the row highlighted in blue indicates our chosen setting.

\noindent\textbf{(a) Single-layer Performance (Table~\ref{tab:subtable1}).} We apply the affinity-based weighted average pooling and token mixing either after/before the projection module (A.\,Proj., B.\,Proj.) or at different layers of the LLM (11,~19-23,~31). Layer20 attains the highest success rates overall. The results show that layers after/before the projection module (A.\,Proj., B.\,Proj.) are not proper locations to apply IVRA. It is important that IVRA is largely layer-agnostic as any layer above 19 yields consistently strong gains.

\noindent\textbf{(b) Multi-layer Performance (Table~\ref{tab:subtable2}).} We evaluate stacking the token mixing in consecutive layers starting from layer~20. Using a single layer (\#~layers\,=\,1) improves the most across all tasks relative to the baseline. Adding more layers shows the similar performance but does not yield large additional gains. Therefore, we select a single layer to strike a balance between performance and additional complexity.

\noindent\textbf{(c) Integration Locations in Transformer Block (Table~\ref{tab:subtable3}).} We compare five positions (P0--P4) within each transformer block: \textbf{P0} (input to the block), \textbf{P1} (after the attention layernorm), \textbf{P2} (output of the attention module), \textbf{P3} (after the attention residual), and \textbf{P4} (after the MLP layernorm). Performing pooling and mixing at \textbf{P0} yields the strongest overall results, suggesting that early injection of affinity cues preserves spatial detail more effectively throughout the subsequent transformations.

\noindent\textbf{(d) Token-Mixing Coefficient (\(\lambda\)) (Table~\ref{tab:subtable4}).} We sweep \(\lambda \in \{1,\,0.7,\,0.3,\,0\}\) to control the convex blend \(v_i^{\mathrm{mix}}=(1-\lambda)\,v_i+\lambda\,v_i'\) between original and affinity‑pooled tokens. A \emph{moderate} mix performs best: \(\lambda=0.3\) attains the highest overall average and the strongest \textbf{NO} (61.3\%), while remaining competitive on \textbf{OP}/\textbf{OC} (73.1\%/70.4\%). Larger \(\lambda\) (0.7–1.0) slightly benefits \textbf{NT} (30\%) but degrades \textbf{NO}/\textbf{OC}, whereas \(\lambda=0\) (no affinity injection) is uniformly worse (22.5–69.6\%). We therefore choose \(\lambda=0.3\) as a balanced setting that avoids over‑smoothing at high \(\lambda\) and under‑utilization at \(\lambda=0\).

\section{Conclusion}
\label{sec:conclusion}
In summary, we presented \textbf{IVRA}, a lightwegiht, training-free inference-time technique for restoring spatial structure in visual-language-action (VLA) models. IVRA injects encoder-derived \emph{affinity hints} into a selected intermediate layer of the language model, reweighting flattened visual tokens using patchwise correlations. This preserves instance-level cues such as object boundaries and attribute relations that are critical for various robot tasks such as precise grasping, placement, and multi-step manipulation. Across both 2D and 3D benchmarks (VIMA and LIBERO) and real-robot tasks, IVRA consistently improves strong VLA baselines, including \llara, OpenVLA, and FLOWER. The results demonstrate the IVRA's practical drop-in enhancement of fine-grained grounding and action generation in multimodal robot policies.

\section*{Reproducibility}
We will publicly release our code. All models used in our paper are publicly available. Additionally, our paper provides detailed descriptions of the methodology, experimental setup, and evaluation protocols to support faithful replication of our results.

\section*{Acknowledgements:}
This research was financially supported by the Ministry of Trade, Industry, and Energy (MOTIE), Korea, under the “Global Industrial Technology Cooperation Center (GITCC) program” supervised by the Korea Institute for Advancement of Technology (KIAT).(Task No. P0028420). This research was supported by the National Research Council of Science \& Technology (NST) grant by the Korea government (MSIT) (No. GTL25041-000).

\bibliographystyle{IEEEtran}
\bibliography{main}

\begin{thebibliography}{10}
\providecommand{\url}[1]{#1}
\csname url@rmstyle\endcsname
\providecommand{\newblock}{\relax}
\providecommand{\bibinfo}[2]{#2}
\providecommand\BIBentrySTDinterwordspacing{\spaceskip=0pt\relax}
\providecommand\BIBentryALTinterwordstretchfactor{4}
\providecommand\BIBentryALTinterwordspacing{\spaceskip=\fontdimen2\font plus
\BIBentryALTinterwordstretchfactor\fontdimen3\font minus \fontdimen4\font\relax}
\providecommand\BIBforeignlanguage[2]{{%
\expandafter\ifx\csname l@#1\endcsname\relax
\typeout{** WARNING: IEEEtran.bst: No hyphenation pattern has been}%
\typeout{** loaded for the language `#1'. Using the pattern for}%
\typeout{** the default language instead.}%
\else
\language=\csname l@#1\endcsname
\fi
#2}}

\bibitem{li2024llara}
X.~Li, C.~Mata, J.~Park, K.~Kahatapitiya, Y.~S. Jang, J.~Shang, K.~Ranasinghe, R.~Burgert, M.~Cai, Y.~J. Lee, and M.~S. Ryoo, ``{LLaRA}: Supercharging robot learning data for vision-language policy,'' \emph{arXiv preprint arXiv:2406.20095}, 2024.

\bibitem{kim2024openvla}
M.~J. Kim, K.~Pertsch, S.~Karamcheti, T.~Xiao, A.~Balakrishna, S.~Nair, R.~Rafailov, E.~Foster, G.~Lam, P.~Sanketi, Q.~Vuong, T.~Kollar, B.~Burchfiel, R.~Tedrake, D.~Sadigh, S.~Levine, P.~Liang, and C.~Finn, ``{OpenVLA}: An open-source vision-language-action model,'' \emph{arXiv preprint arXiv:2406.09246}, 2024.

\bibitem{reuss2025flower}
M.~Reuss, H.~Zhou, M.~R{\"u}hle, {\"O}.~E. Ya{\u{g}}murlu, F.~Otto, and R.~Lioutikov, ``Flower: Democratizing generalist robot policies with efficient vision-language-action flow policies,'' \emph{arXiv preprint arXiv:2509.04996}, 2025.

\bibitem{niu2024llarva}
D.~Niu, Y.~Sharma, G.~Biamby, J.~Quenum, Y.~Bai, B.~Shi, T.~Darrell, and R.~Herzig, ``{LLARVA}: Vision-action instruction tuning enhances robot learning,'' \emph{arXiv preprint arXiv:2406.11815}, 2024.

\bibitem{radford2021clip}
A.~Radford, J.~W. Kim, C.~Hallacy, A.~Ramesh, G.~Goh, S.~Agarwal, G.~Sastry, A.~Askell, P.~Mishkin, J.~Clark, G.~Krueger, and I.~Sutskever, ``Learning transferable visual models from natural language supervision,'' in \emph{Proceedings of the 38th International Conference on Machine Learning}, 2021.

\bibitem{caron2021dino}
M.~Caron, H.~Touvron, I.~Misra, H.~J{\'e}gou, J.~Mairal, P.~Bojanowski, and A.~Joulin, ``Emerging properties in self-supervised vision transformers,'' in \emph{Proceedings of the IEEE International Conference on Computer Vision}, 2021.

\bibitem{pantazopoulos2024lost}
G.~Pantazopoulos, A.~Suglia, O.~Lemon, and A.~Eshghi, ``Lost in space: Probing fine-grained spatial understanding in vision and language resamplers,'' \emph{Proceedings of the 2024 NAACL-HLT (Volume 2: Short Papers)}, pp. 540--549, 2024.

\bibitem{pmlr-v164-shridhar22a}
M.~Shridhar, L.~Manuelli, and D.~Fox, ``Cliport: What and where pathways for robotic manipulation,'' in \emph{Proceedings of the 5th Conference on Robot Learning (CoRL)}, ser. Proceedings of Machine Learning Research, A.~Faust, D.~Hsu, and G.~Neumann, Eds., vol. 164.\hskip 1em plus 0.5em minus 0.4em\relax PMLR, August 2022, pp. 894--906.

\bibitem{Ranasinghe2025PixelMA}
K.~Ranasinghe, X.~Li, C.~Mata, J.~S. Park, and M.~S. Ryoo, ``Pixel motion as universal representation for robot control,'' \emph{ArXiv}, 2025.

\bibitem{han2025space}
B.~Han, W.-h. Yun, B.-S. Seo, and J.~Kim, ``Space-aware instruction tuning: Dataset and benchmark for guide dog robots assisting the visually impaired,'' \emph{arXiv preprint arXiv:2502.07183}, 2025.

\bibitem{jiang2023vima}
Y.~Jiang, A.~Gupta, Z.~Zhang, G.~Wang, Y.~Dou, Y.~Chen, L.~Fei-Fei, A.~Anandkumar, Y.~Zhu, and L.~Fan, ``Vima: General robot manipulation with multimodal prompts,'' in \emph{ICML}, 2023.

\bibitem{alayrac2022flamingo}
J.~Alayrac, J.~Donahue, P.~Luc, A.~Miech, I.~Barr, Y.~Hasson, K.~Lenc, A.~Mensch, K.~Millican, M.~Reynolds, R.~Ring, E.~Rutherford, S.~Cabi, T.~Han, Z.~Gong, S.~Samangooei, M.~Monteiro, J.~Menick, S.~Borgeaud, A.~Brock, A.~Nematzadeh, S.~Sharifzadeh, M.~Binkowski, R.~Barreira, O.~Vinyals, A.~Zisserman, and K.~Simonyan, ``Flamingo: a visual language model for few-shot learning,'' in \emph{Advances in Neural Information Processing Systems 35 (NeurIPS 2022)}, 2022.

\bibitem{li2023blip2}
J.~Li, D.~Li, S.~Savarese, and S.~C.~H. Hoi, ``{BLIP-2}: Bootstrapping language-image pre-training with frozen image encoders and large language models,'' in \emph{Proceedings of the 40th International Conference on Machine Learning (ICML)}, ser. PMLR, vol. 202, 2023, pp. 19\,730--19\,742.

\bibitem{liu2023llava}
H.~Liu, C.~Li, Q.~Wu, and Y.~J. Lee, ``Visual instruction tuning,'' in \emph{NeurIPS}, 2023.

\bibitem{jiang2023clipdino}
D.~Jiang, Y.~Liu, S.~Liu, X.~Zhang, J.~Li, H.~Xiong, and Q.~Tian, ``From {CLIP} to {DINO}: Visual encoders shout in multi-modal large language models,'' \emph{arXiv preprint arXiv:2310.08825}, 2023.

\bibitem{Zhong_2022_CVPR}
Y.~Zhong, J.~Yang, P.~Zhang, C.~Li, N.~F. Codella, L.~H. Li, L.~Zhou, X.~Dai, L.~Yuan, Y.~Li, and J.~Gao, ``Regionclip: Region-based language-image pretraining,'' in \emph{Proceedings of the IEEE/CVF Conference on Computer Vision and Pattern Recognition (CVPR)}, 2022, pp. 16\,793--16\,803.

\bibitem{Guo_2024_CVPR}
Q.~Guo, S.~De~Mello, H.~Yin, W.~Byeon, K.~C. Cheung, Y.~Yu, P.~Luo, and S.~Liu, ``Regiongpt: Towards region understanding vision language model,'' in \emph{Proceedings of the IEEE/CVF Conference on Computer Vision and Pattern Recognition (CVPR)}, 2024, pp. 13\,796--13\,806.

\bibitem{Ran2024LearningTL}
K.~Ranasinghe, S.~N. Shukla, O.~Poursaeed, M.~S. Ryoo, and T.-Y. Lin, ``Learning to localize objects improves spatial reasoning in visual-llms,'' \emph{2024 IEEE/CVF Conference on Computer Vision and Pattern Recognition (CVPR)}, 2024.

\bibitem{Liu_2024_ECCV}
S.~Liu, Z.~Zeng, T.~Ren, F.~Li, H.~Zhang, J.~Yang, Q.~Jiang, C.~Li, J.~Yang, H.~Su, J.~Zhu, and L.~Zhang, ``Grounding dino: Marrying dino with grounded pre-training for open-set object detection,'' in \emph{Proceedings of the 18th European Conference on Computer Vision (ECCV)}, 2024.

\bibitem{zhou2024hints}
H.~Zhou, Z.~Gao, M.~Ye, Z.~Chen, T.~Cao, and H.~Qi, ``Hints of prompt: Enhancing visual representation for multimodal llms in autonomous driving,'' \emph{arXiv preprint arXiv:2411.13076}, 2024.

\bibitem{wysoczanska2024clipdiy}
M.~Wysocza\'{n}ska, M.~Ramamonjisoa, T.~Trzci\'{n}ski, and O.~Sim\'{e}oni, ``{CLIP-DIY}: {CLIP} dense inference yields open-vocabulary semantic segmentation for-free,'' in \emph{Proceedings of the IEEE/CVF Winter Conference on Applications of Computer Vision (WACV)}, 2024, pp. 1392--1402.

\bibitem{Cai_2024_CVPR}
M.~Cai, H.~Liu, S.~K. Mustikovela, G.~P. Meyer, Y.~Chai, D.~Park, and Y.~J. Lee, ``Vip-llava: Making large multimodal models understand arbitrary visual prompts,'' in \emph{Proceedings of the IEEE/CVF Conference on Computer Vision and Pattern Recognition (CVPR)}, 2024, pp. 12\,914--12\,923.

\bibitem{Xu_2022_CVPR}
J.~Xu, S.~De~Mello, S.~Liu, W.~Byeon, T.~Breuel, J.~Kautz, and X.~Wang, ``Groupvit: Semantic segmentation emerges from text supervision,'' in \emph{Proceedings of the IEEE/CVF Conference on Computer Vision and Pattern Recognition (CVPR)}, 2022, pp. 18\,134--18\,144.

\bibitem{Liang_2023_CVPR}
F.~Liang, B.~Wu, X.~Dai, K.~Li, Y.~Zhao, H.~Zhang, P.~Zhang, P.~Vajda, and D.~Marculescu, ``Open-vocabulary semantic segmentation with mask-adapted clip,'' in \emph{Proceedings of the IEEE/CVF Conference on Computer Vision and Pattern Recognition (CVPR)}, 2023, pp. 7061--7070.

\bibitem{Ran2022PerceptualGI}
K.~Ranasinghe, B.~McKinzie, S.~Ravi, Y.~Yang, A.~Toshev, and J.~Shlens, ``Perceptual grouping in contrastive vision-language models,'' \emph{2023 IEEE/CVF International Conference on Computer Vision (ICCV)}, 2022.

\bibitem{Shridhar_2022_CoRL}
M.~Shridhar, L.~Manuelli, and D.~Fox, ``Cliport: What and where pathways for robotic manipulation,'' in \emph{Proceedings of The 6th Conference on Robot Learning (CoRL)}, 2022, pp. 894--906.

\bibitem{Driess_2023_ICML}
D.~Driess, F.~Xia, M.~S.~M. Sajjadi, C.~Lynch, A.~Chowdhery, B.~Ichter, A.~Wahid, J.~Tompson, Q.~Vuong, T.~Yu, W.~Huang, Y.~Chebotar, P.~Sermanet, D.~Duckworth, S.~Levine, V.~Vanhoucke, K.~Hausman, M.~Toussaint, K.~Greff, A.~Zeng, I.~Mordatch, and P.~Florence, ``Palm-e: An embodied multimodal language model,'' in \emph{Proceedings of the 40th International Conference on Machine Learning (ICML)}, 2023, pp. 8469--8488.

\bibitem{chi2023diffusion}
C.~Chi, S.~Feng, Y.~Du, Z.~Xu, E.~Cousineau, B.~Burchfiel, and S.~Song, ``Diffusion policy: Visuomotor policy learning via action diffusion,'' \emph{Robotics science and systems (RSS)}, 2023.

\bibitem{octo_2023}
{Octo Model Team}, D.~Ghosh, H.~Walke, K.~Pertsch, K.~Black, O.~Mees, S.~Dasari, J.~Hejna, C.~Xu, J.~Luo, T.~Kreiman, Y.~Tan, L.~Y. Chen, P.~Sanketi, Q.~Vuong, T.~Xiao, D.~Sadigh, C.~Finn, and S.~Levine, ``Octo: An open-source generalist robot policy,'' in \emph{Robotics science and systems (RSS)}, Delft, Netherlands, 2024.

\bibitem{brohan2023rt}
A.~Brohan, N.~Brown, J.~Carbajal, Y.~Chebotar, X.~Chen, K.~Choromanski, T.~Ding, D.~Driess, A.~Dubey, C.~Finn, \emph{et~al.}, ``Rt-2: Vision-language-action models transfer web knowledge to robotic control,'' \emph{arXiv preprint arXiv:2307.15818}, 2023.

\bibitem{wysoczanska2024clipdinoiser}
M.~Wysocza{\'n}ska, O.~Sim{\'e}oni, M.~Ramamonjisoa, A.~Bursuc, T.~Trzci{\'n}ski, and P.~P{\'e}rez, ``Clip-dinoiser: Teaching clip a few dino tricks for open-vocabulary semantic segmentation,'' in \emph{European Conference on Computer Vision}.\hskip 1em plus 0.5em minus 0.4em\relax Springer, 2024, pp. 320--337.

\end{thebibliography}

\end{document}